\documentclass[a4paper,fleqn]{cas-sc}

\usepackage[numbers,longnamesfirst]{natbib}

\usepackage{algorithm, algorithmic}
\usepackage{makecell}
\usepackage{float}
\usepackage{amsmath}
\def\tsc#1{\csdef{#1}{\textsc{\lowercase{#1}}\xspace}}
\tsc{WGM}
\tsc{QE}
\tsc{EP}
\tsc{PMS}
\tsc{BEC}
\tsc{DE}
\begin{document}
\let\WriteBookmarks\relax
\def\floatpagepagefraction{1}
\def\textpagefraction{.001}
\shorttitle{Similarity Based Stratified Splitting}
\shortauthors{FC Farias, TB Ludermir, CJA Bastos-Filho.}
%\begin{frontmatter}

\title [mode = title]{Similarity Based Stratified Splitting: an approach to train better classifiers}
% \tnotemark[1,2]

% \tnotetext[1]{This document is the results of the research
%   project funded by the National Science Foundation.}

% \tnotetext[2]{The second title footnote which is a longer text matter
%   to fill through the whole text width and overflow into
%   another line in the footnotes area of the first page.}

\author[1,2]{Felipe C. Farias}[orcid=0000-0001-7411-5562]

% \cormark[1]
\ead{felipefariax@gmail.com}

\author[2]{Teresa B. Ludermir}[orcid=0000-0002-8980-6742]
\ead{tbl@cin.ufpe.br}

\author[3]{Carmelo J. A. Bastos-Filho}[orcid=0000-0002-0924-5341]
\ead{carmelofilho@ieee.org}

\address[1]{Universidade Federal de Pernambuco, Centro de Informática, Recife PE, Brasil}
\address[2]{Instituto Federal de Educação, Ciência e Tecnologia de Pernambuco, Paulista PE, Brasil}
\address[3]{Universidade de Pernambuco, EComp, Recife PE, Brasil}

\begin{abstract}
We propose a Similarity-Based Stratified Splitting (SBSS) technique, which uses both the output and input space information to split the data. The splits are generated using similarity functions among samples to place similar samples in different splits. This approach allows for a better representation of the data in the training phase. This strategy leads to a more realistic performance estimation when used in real-world applications. We evaluate our proposal in twenty-two benchmark datasets with classifiers such as Multi-Layer Perceptron, Support Vector Machine, Random Forest and K-Nearest Neighbors, and five similarity functions Cityblock, Chebyshev, Cosine, Correlation, and Euclidean. According to the Wilcoxon Sign-Rank test, our approach consistently outperformed ordinary stratified 10-fold cross-validation in 75\% of the assessed scenarios.
\end{abstract}

% \includegraphics{figs/grabs.pdf}

% Highlights should be submitted in a separate editable file in the online submission system. Please use 'Highlights' in the file name and include 3 to 5 bullet points (maximum 85 characters, including spaces, per bullet point).
\begin{highlights}
\item Split the data using input and output distribution information;
\item An approach to train better classifiers;
\item Classifier agnostic and low-cost proposal to increase models' performance.
\end{highlights}

% maximo 6 keywords
\begin{keywords}
k-fold \sep cross-validation \sep machine learning \sep samples similarity \sep classifiers \sep data splitting
\end{keywords}

\maketitle

\section{Introduction}
Machine Learning systems use data to extract knowledge. The goal is to store information in the internal parameters to analyze future unseen data. Some learning paradigms have been used in machine learning. In Supervised learning, the inputs and desired outputs (targets) are presented to the model. In Unsupervised learning, no output is given—the model clusters data based on the similarity among the elements. Conversely, in the Reinforcement learning paradigm, the model acts in an environment and evaluates if the last actions led to a better environment metric, receiving a reward or a punishment. Modeling a supervised learning system requires a dataset $D = \{(x_i, y_i) | x_i \in X, y_i \in Y\}$, where $X$ is the input in the feature space, and $Y$ is the output, which may be labeled for classification or real values for regression applications.

We may define a Machine Learning system as a combination of three elements: (i) a model $M$, which is a mathematical function mapping the domain $X$ into image $Y$, processing the input $X$ using its internal parameters $W$; (ii) an Error/Cost/Loss Function $L=error(M(X), Y)$, which evaluates the performance of $M$ in $D$; and (iii) an Optimizer $(O)$, which minimizes the function $L$. The training phase of a Learning System tries to find the best $ W $ for a model $ M $ using data $ D $ to estimate the parameters $ W $, minimizing the error $ L $ using $ O $. In other words, $W = argmin(L(M(X), Y))$ with $W$ estimated by $O(L)$.

A typical approach to conceive a Supervised Learning System needs to have data $X_{train}$ (training set) presented to the model to estimate the parameters $W$ and different data $X_{test}$ (validation set or test set), which is not used to modify the models' parameters $W$, but to evaluate its performance, serving as an estimation of real-world data during the inference phase. Several strategies have been developed to split data better~\cite{kohavi1995study}. The error considering the training set should never be used alone as a model's performance estimator since some problems during the training phase may arise. The most common problems are over-fitting and under-fitting~\cite{Bishop2006,russell2020artificial}. To mitigate these problems, we can split the data into two or more subsets. We briefly discuss some split data methods in the next paragraphs.

%\subsection{Holdout}
One of the most common split data methods, probably due to its simplicity, is the Holdout Splitting. Holdout randomly divides the original dataset into a training set from which the algorithm produces the model $ M $ and a test set on which the performance of $ M $ is evaluated \cite{russell2020artificial}. A common choice is to use $75\%$ of the samples to the training set and $25\%$ to the test set. It is desired that the training and test sets contain different samples and follow approximately the same distribution, which is not always the case.

%\subsection{K-Fold cross-validation}
In the K-Fold cross-validation, the data is divided into $k$ subsets. Then $k$ rounds of learning are performed. On each round, $1/k$ of the data is used as the test set, while the remaining samples are used as the training data. The average test set performance of the $k$ rounds should be calculated. Popular values for $k$ are 5 and 10 \cite{Breiman1992,wong2015performancekfold,russell2020artificial}. An interesting analysis of K-Fold cross-validation estimates can be found  in~\cite{Wong2017,Jung2018,Wong2019}.

%\subsection{Leave One Out}
Leave One Out is K-Fold cross-validation when $k$ assumes the original dataset number of samples~\cite{russell2020artificial}. Consequently, it uses a test set with just one sample and the training set with all others. The process is repeated until every single sample belongs to the test set once. Then, the model's performances in the test sets are also averaged.

If the dataset is imbalanced, a stratified splitting may be recommended since it divides the dataset, maintaining the class proportion in each of the subsets created. It is even worse when there is a small number of samples to be trained. 

Cross-validation is used to estimate the model generalization to an independent dataset. It is commonly applied to learning systems to predict its future performance or how accurate it would be when used in the real world, where the model never received the data during the training phase. It can also be used to (i) stop the training phase at a point which if the model was trained below or above that quantity, and it could present under-fitting or over-fitting behaviors, respectively; (ii) compare the performance of different models submitted to the same data; (iii) select the best model over several runs when there is a stochastic component intrinsic to the model; (iv) choose the classifiers that will compose an ensemble system.

Since it is a common approach to use cross-validation to obtain the best model to run on real-world problems, we should expect that our data during the training phase could have approximately the same distribution of data received in real-world applications. Besides, several algorithms rely on the fact that the samples in each subset are Independent and Identically Distributed \cite{Haykin2008,japkowicz2011evaluating}. These are reasonable assumptions since the process of learning means to model the distribution of the training data as close as possible, expecting that future data seen in the real-world follow the approximated distribution. None of those mentioned earlier strategies grants it or even have robust mechanisms to partially induce this assumption since they use the labels or the output distribution to split the data, ignoring all the input space distribution information. The DUPLEX algorithm~\cite{snee1977validation} uses the input space information to divide the data into two disjoint sets with statistical similarity and cover almost the same region of input space. It uses the Euclidean distance between all pairs of samples to place the most distant samples in the same set, alternating from the train set and test set. As it was created to solve regression problems, it does not consider the output distribution of the labels nor the stratification process.

Our research question is posed as follows:
\begin{center}
\fbox{%
  \parbox[l]{0.95\textwidth}{
    \textbf{RQ:} Can we use input and output space information during the data splitting process to cover all the regions where samples exist, aiming to maintain approximately the same statistical properties, and to generate better data to create better classifiers?
  }%
}    
\end{center}

Consider a real-world intelligent system. The most common approach to train an Artificial Neural Network in this task is to (i) split the data into train and validation sets; (ii) define an upper bound limit of epochs for the model to be trained, and (iii) define a patience threshold for early stopping the training before the limit of epochs based on how many times the model presented consecutive performance decreases. At the end of each epoch, considering the training set, the validation set is used to assess the model's performance. As the validation set is not used to change the model's parameters, it acts as a proxy set of the data that the model will be presented in real-world operation. It is natural to expect that higher performances in the validation set consequently generate better models in real-world usage, leading us to use the model that shows the smallest error in the validation set. In order to happen what we expect, it is a necessary condition that the validation set has samples drawn from approximately the same distribution of the real-world data that the model will see in the future. In other words, our splitting process should look not only to the output space but also the input distribution of samples plays an essential role in the model's performance. In this work, we propose a low-cost method to increase the model's performance by better selecting the training data to be presented using similarity functions to place similar samples in different splits.

We organized the remainder of this work as follows. We present the background information, describe the proposed methodology to split the data using similarity functions, and provide experiment details in Section 2. We show the results and discussions in Section 3. Finally, we present conclusions and future works in Section 4.

\section{Material and Methods}
\subsection{Our Proposal}

In our proposal, we consider the output space and the distribution of inputs/features space to create the splits to validate our hypothesis. We call it Similarity-Based Stratified Splitting (SBSS). Even though our technique is not limited to a specific data splitting strategy, we focused our evaluations on Stratified 10-Fold. We used its folds to compose the train and test set, the latest mimicking the model's real-world usage.

\begin{figure}[hbt!]
    \centering
    \includegraphics[width=0.5\textwidth]{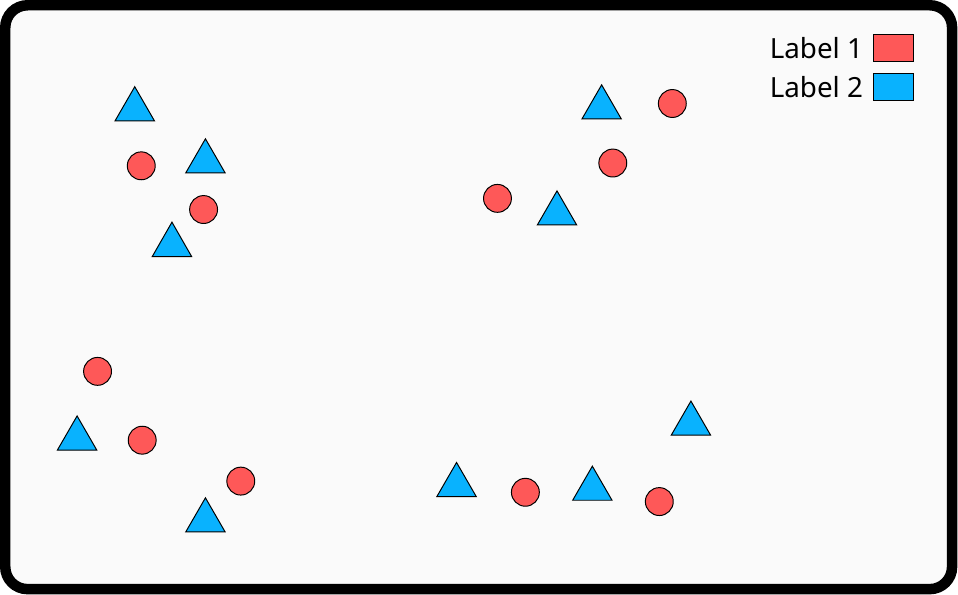}
    \caption{Example of dataset with 2 labels.} \label{fig:dataset}
\end{figure}

To illustrate our hypothesis, in Figure~\ref{fig:dataset}, we show a dataset with 20 samples divided into two labels.

\begin{figure}[hbt!]
    \centering
    \includegraphics[width=0.5\textwidth]{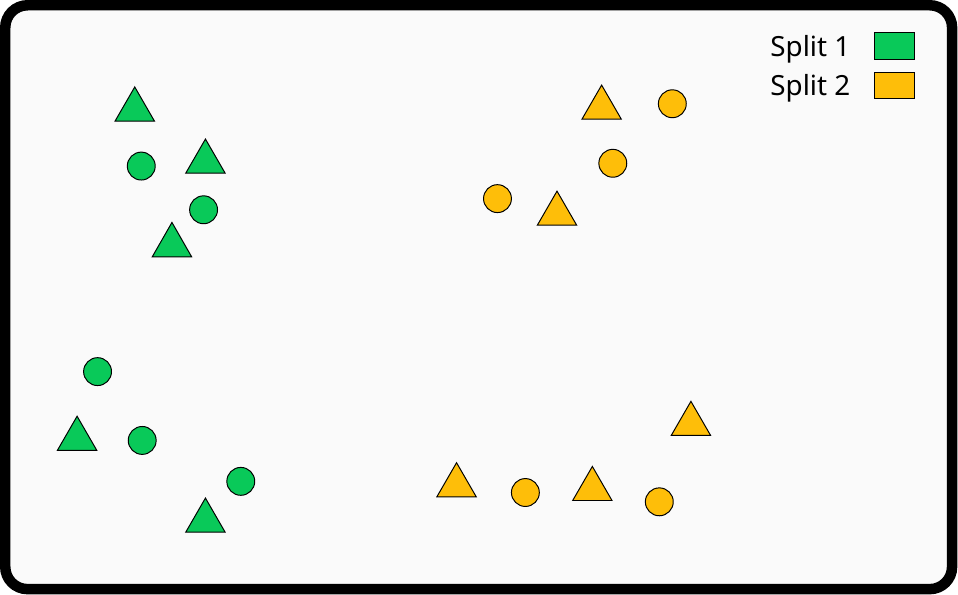}
    \caption{Stratified splits ignoring the feature space.} \label{fig:bad_split}
\end{figure}

If we use a conventional stratified split process that does not consider the input space distribution, at worst case, we can create two bad splits, as shown in Figure~\ref{fig:bad_split}. If we use Split 1 as the training set and Split 2 as the test set, the model will probably perform poorly since the samples lie in a region that the model was not exposed to data and did not learn how to separate the samples at this sub-space. The splitting process is also used to select the hyper-parameters of a model in a proxy subset of the data not presented in the training phase. This proxy subset should follow approximately the same distribution of the real data that the model will be presented when inferring in production. If the samples are not carefully chosen, we may not have consistent performance metrics compared to the real-world deployed scenario, even when the validation/test set presents high-performance measurements.

\begin{figure}[hbt!]
    \centering
    \includegraphics[width=0.5\textwidth]{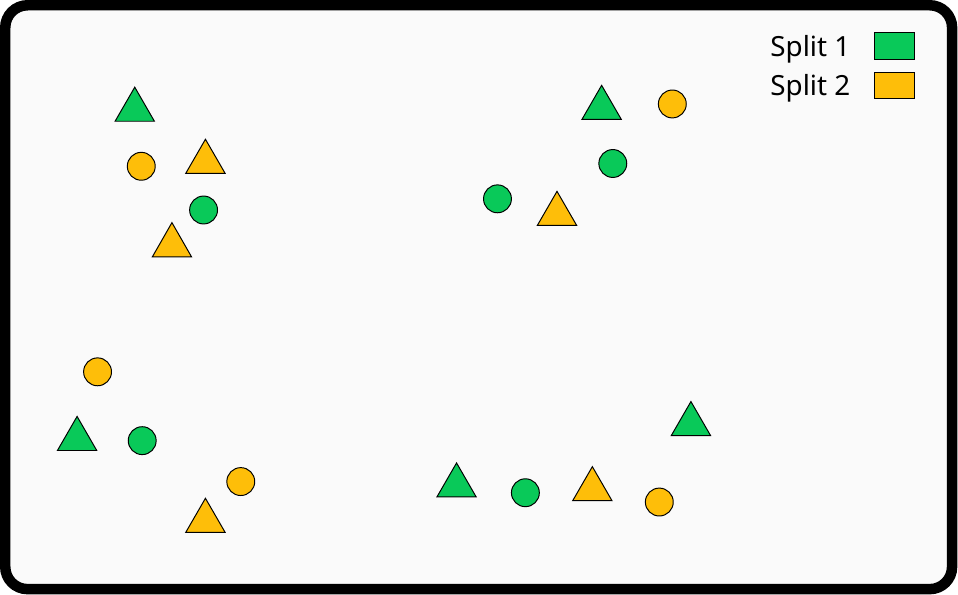}
    \caption{One expected splitting scenario of our approach.} \label{fig:sbsf}
\end{figure}

We try to place similar elements belonging to the same label in different splits to maintain the input and output distribution over all the splits approximately equal. First, we create N splits and we calculate the similarity matrix of samples' features. For each label, we find the pivot sample (the sample with the largest similarity to all other samples of the same label) and the next N-1 most similar samples. Then we shuffle these picked samples to guarantee the stochastic behavior and append each sample to each split. After, we remove the picked samples and repeat the process until there is no sample left in the dataset. This process is summarized in Algorithm~\ref{alg:sbsf}. The Python code is also available in \footnote{https://github.com/felipefariax/sbss}. In Figure~\ref{fig:sbsf}, we have an expected scenario of our approach outcome. The distributions of both splits contain more relevant data/information to be learned by the model than the ones in Figure~\ref{fig:bad_split}.

One could argue that this approach has no value since we are not randomly picking samples and the real-world usage of the model the outcomes are random; therefore, we could not assume that it will follow the same distribution. We believe that if this happens remarkably, probably the dataset does not represent the task's population. In this case, more data should be collected before start modeling in order to represent the task to be learned better. Once Learning Systems recognizes patterns in the data, it will probably fail to recognize these patterns when using the real-world model if it does not have similar patterns to learn. For example, an application of object recognition trained with only one color of a specific object may have difficulties to detect colored objects of the same nature and vice-versa. The model would probably perform better if the same ratios of one color and colored objects are presented during the training and validation phase. As this information may not be explicit, one could use the similarities between the objects' colors to drive this behavior. Also, perhaps the real-world application is not generating random outcomes. The samples can be drawn from a distribution that classical statistical probability distributions have difficulties handling or representing graphically. It could give us the notion of distribution that we would like them to have.

%For example, an application of face detection trained with only white faces may have difficulties to detect black faces and vice-versa. The model would probably perform better if the same ratio of black and white faces are presented during the training and validation phase. As this information may not be explicit, one could use the similarities between the faces in order to drive this behavior (ALGUM EXEMPLO MELHOR? NAO SEI SE ISSO PODE GERAR DISCUSSAO POR BLACK LIVES MATTER E DESVIAR O FOCO...). Also, perhaps the real world application is not generating random outcomes. The samples can be drawn from a distribution that classical statistical probability distributions have difficulties handling and/or representing graphically that could give us the notion of distribution that we would like them to have.

\begin{algorithm}[hbt!]
 \caption{Algorithm for Similarity Based Stratified K-Fold Splitting}\label{alg:sbsf}
 \begin{algorithmic}[1]
 \REQUIRE X, Y, k // Normalized Inputs, Outputs, number of splits
%  \\ \textit{Initialisation} :
%   \STATE first statement
%  \\ \textit{LOOP Process}
    \STATE splits = [[]*k] // Create a list with K sublists
    \STATE split\_labels = []
    
    \STATE similarity\_matrix = similarity(X, X) // Calculate the similarity between all the samples
    
    \FOR {label in labels:}
        \WHILE {exist\_samples\_dataset(X, Y, label)}
            \STATE //Index of the most similar sample to all samples not already picked of this label. i.e. largest similarity sum to other samples of the same label
            \STATE pivot\_idx = get\_pivot\_sample\_idx(X, Y, label)
            \item[]
    
            \STATE picked\_samples.append(pivot\_idx)
            \item[]
            
            \STATE //Get the next N-1 most similar samples
            \FOR {split in N-1:}
                \STATE closest\_ix = get\_most\_similar\_sample(X, Y, label, picked\_samples)
                \STATE picked\_samples.append(closest\_ix)
            \ENDFOR
            % \STATE // Get the k-1 indexes of the most similars not already picked samples to the pivot\_idx sample belonging to this label.
            % \STATE most\_similar\_idxs = get\_n\_picked\_samples\_idxs(X, Y, label, pivot\_idx, k-1) 
            
            \item[]
            
            % \STATE picked\_idxs = [pivot\_idx] + most\_similar\_idxs
            \STATE shuffle(picked\_samples)
            
            \item[]
            \STATE // Filling splits
            \FOR {i = 0...k-1}
                \STATE splits[i].append(picked\_samples[i])
                \STATE split\_labels.append(label) 
            \ENDFOR
            
            \item[]
            
            \STATE X, Y = remove\_samples\_from\_dataset(X, Y, picked\_samples)
        \ENDWHILE
    \ENDFOR
    \item[]
\STATE // Contains the index of each sample belonging to each split as a ``k'' by ``nb\_samples'' matrix. Each col has indexes of the samples belonging to approximately the same region in the same label regarding the original dataset.
 \RETURN splits, split\_labels
 
 \end{algorithmic} 
 \end{algorithm}
 
Note that any similarity/distance function can be used. We have assessed five similarity functions, namely (i) Cityblock, (ii) Chebyshev, (iii) Euclidean, (iv) Cosine, (v) Correlation to show this. The equations of each function are presented in Eq.~\ref{eq:chebyshev}-\ref{eq:correlation}.

\begin{align}
    \label{eq:chebyshev}
    Chebyshev & = max|u_i-v_i| \\
    Cityblock & = \sum_{i}{|u_i-v_i|} \\
    Euclidean & = \sqrt{\sum_{i}{(u_i-v_i)^2}} \\
    Cosine & = 1 - \frac{u \cdot v}{\left\|u\right\|_2 \left\|v\right\|_2} \\
    \label{eq:correlation}
    Correlation & = 1 - \frac{(u-\Bar{u}) \cdot (v-\Bar{v})}{\left\|(u-\Bar{u})\right\|_2 \left\|(v-\Bar{v})\right\|_2}
\end{align}

where $u$ and $v$ are two input vectors; $i$ is the dimension index; $|\cdot|$ is the absolute value and $\left\|\cdot\right\|_2$ refers to a L2-norm.

\subsection{Classifiers}
We have assessed different families of classifiers to show the transversality of our proposal. We deployed the following algorithms: (i) a K-Nearest Neighbors (KNN)~\cite{cover1967nearest} (ii) Random Forest (RF) \cite{breiman2001random} classifier with 100 trees; (iii) Support Vector Machine (SVM) \cite{Cortes1995} and (iv) Multilayer Perceptron (MLP) \cite{Haykin2008} with 20 hidden neurons without a validation set to early stop the model, 300 epochs and 0.001 as the learning rate. All the other hyperparameters were used with the default values available in the Scikit-Learn \cite{pedregosa2011scikit} Python library. As we intend to show that SBSS can improve the classifiers' performance, we have not done much hyperparameter experimentation, meaning that we could achieve even better performances if the hyperparameters were carefully chosen. We have chosen these classifiers due to their different nature bases such as instance-based, decision tree, hyperplane separation, and regression-based methodologies.

\subsection{Datasets}
We assessed the proposed algorithm in several situations, such as many features and labels, a low number of samples, and dataset imbalance. We have used 22 datasets from UCI~\cite{UCI} presented in Table~\ref{table:datasets}.
We calculated the Imbalance of each dataset by adapting the suggestion in \cite{imbalancemetric} according to Eq.~\ref{eq:dataset_imbalance}, resulting in 0 when the dataset is balanced and 1 otherwise. 

\begin{equation}
    \label{eq:dataset_imbalance}
    1 - \frac{\sum_{i=1}^{k}\frac{c_i}{n}log(\frac{c_i}{n})}{log(k)}
\end{equation}
where $n$ is the number of samples; $k$ is the number of labels, and $c_i$ is the number of samples in label $i$.

\begin{table}[hbt!]
    \caption{Datasets used as benchmarks}\label{table:datasets}
    \begin{center}
        \begin{tabular}{p{6cm}cccc}
            \toprule
            \textbf{Dataset}                 & \textbf{\# Features} & \textbf{\# Labels} & \textbf{\# Samples} & \textbf{Imbalance} \\
            \midrule
            balance-scale                    & 4                    & 3                  & 625                 & 0.17               \\
            blood-transfusion-service-center (btsc) & 4                    & 2                  & 748                 & 0.21               \\
            car                              & 6                    & 4                  & 1728                & 0.40               \\
            diabetes                         & 8                    & 2                  & 768                 & 0.07               \\
            tic-tac-toe                      & 9                    & 2                  & 958                 & 0.07               \\
            ilpd                             & 10                   & 2                  & 583                 & 0.14               \\
            vowel                            & 12                   & 11                 & 990                 & 0.00               \\
            australian                       & 14                   & 2                  & 690                 & 0.01               \\
            climate-model-simulation-crashes (cmsc) & 18                   & 2                  & 540                 & 0.58               \\
            vehicle                          & 18                   & 4                  & 846                 & 0.00               \\
            credit-g                         & 20                   & 2                  & 1000                & 0.12               \\
            wdbc                             & 30                   & 2                  & 569                 & 0.05               \\
            ionosphere                       & 34                   & 2                  & 351                 & 0.06               \\
            satimage                         & 36                   & 6                  & 6430                & 0.04               \\
            libras move                     & 90                   & 15                 & 360                 & 0.00               \\
            hill-valley                      & 100                  & 2                  & 1212                & 0.00               \\
            musk                             & 167                  & 2                  & 6598                & 0.38               \\
            lsvt                             & 310                  & 2                  & 126                 & 0.08               \\
            madelon                          & 500                  & 2                  & 2600                & 0.00               \\
            cnae-9                           & 856                  & 9                  & 1080                & 0.00               \\
            dbworld-bodies                   & 4702                 & 2                  & 64                  & 0.01               \\
            arcene                           & 10000                & 2                  & 200                 & 0.01               \\
            \bottomrule
        \end{tabular}
    \end{center}
\end{table}

\subsection{Experiments}
% we have done $22*5*4*10*10=44000$ SBSF and $22*4*10*10=10000$ ordinary 10-fold simulations.
We have simulated ten experiments applying SBSS to 10-Fold cross-validation, which we called Similarity-Based Stratified 10-Fold (SBSF), totaling 100 executions -- 10 simulations of 10 splits, each split being used as the testing set once. As we have 22 datasets, five similarity measures, and four classifiers, we have done 44,000 SBSF and 10,000 ordinary 10-fold simulations. Although several metrics \cite{Seliya2009} can be used, we have used the average accuracy of the 10-fold averaged splits to compare our approach against the original stratified 10-fold splitting. We have applied the Wilcoxon Signed-Rank test to assess if our approach significantly increases the classifier's performance with $\alpha=0.05$. We use nine folds to compose the train set and one fold as the test set.

\section{Results and Discussions}
In this section, we evaluate and discuss the results of our experiments briefly, comparing the scenarios with and without applying the SBSF split strategy.

We present in Table~\ref{table:similarities_accuracies} the training and test set average accuracy. We present average accuracy, and the average standard deviation of 10 evaluations of 10-fold applied to SBSF and original 10-fold stratified splitting inside the parenthesis. We also show the averaged accuracy difference between SBSF and 10-fold. We can see that the Correlation similarity yields the best test accuracy, while the Euclidean had the worst accuracy among the similarity functions used in the SBSF strategy, even though it is still more significant than the 10-fold strategy. Also, the accuracy increase in the test set was more prominent than in the training set, probably indicating a better generalization of the model since it tends not to have high-density regions that would give more importance due to a bad data splitting. The standard deviation was also reduced in SBSF.

\begin{table}[h!bt]
    \caption{Similarity accuracy averaged over all datasets and models.}\label{table:similarities_accuracies}
    \begin{tabular*}{\tblwidth}{@{} LLLLLLL@{} }
    \toprule
    % Col 1 & Col 2 & Col 3 & Col4 & 5 & 6 & 7\\
     \multirow{3}{*}{\centering\textbf{Similarity}} & \multicolumn{3}{c}{\textbf{Train}} & \multicolumn{3}{c}{\textbf{Test}} \\
     \cline{2-4} \cline{5-7} & \textbf{10-fold} & \textbf{SBSF} & \textbf{\makecell{Difference\\(SBSF-10-fold)}} & \textbf{10-fold} & \textbf{SBSF} & \textbf{\makecell{Difference\\(SBSF-10-fold)}} \\
    \midrule
    Chebyshev   & \makecell{90.129\\(1.16)}            & \makecell{90.492\\(1.104)}        & \makecell{0.363\\(-0.056)} & \makecell{82.027\\(4.326)}           & \makecell{83.050\\(3.758)}        & \makecell{1.023\\(-0.568)} \\
    Cityblock   & \makecell{90.129\\(1.16)}          & \makecell{90.378\\(1.085)}        & \makecell{0.249\\(-0.075)}     & \makecell{82.027\\(4.326)}           & \makecell{83.086\\(3.644)}        & \makecell{1.059\\(-0.682)}      \\
    Euclidean   & \makecell{90.129\\(1.16)}           & \makecell{90.385\\(1.104)}        & \makecell{0.256\\(-0.056)}    & \makecell{82.027\\(4.326)}           & \makecell{82.988\\(3.678)}        & \makecell{0.961\\(-0.648)}     \\
    Cosine      & \makecell{90.129\\(1.16)}           & \makecell{90.380\\(1.102)}        & \makecell{0.251\\(-0.058)}    & \makecell{82.027\\(4.326)}           & \makecell{83.188\\(3.440)}        & \makecell{1.161\\(-0.886)}     \\
    Correlation & \makecell{90.129\\(1.16)}            & \textbf{\makecell{90.536\\(1.101)}}       & \makecell{0.407\\(-0.059)} & \makecell{82.027\\(4.326)}           & \textbf{\makecell{83.363\\(3.574)}}       & \makecell{1.336\\(-0.752)}     \\
    \midrule
    Average & \makecell{90.129\\(1.16)} & \makecell{90.434\\(1.099)} & \makecell{0.305\\(-0.061)} & \makecell{82.027\\(4.326)} & \makecell{83.135\\(3.619)} & \makecell{1.108\\(-0.707)} \\
    \bottomrule
    \end{tabular*}
\end{table}

\begin{figure}[hbt!]
	\centering
		\includegraphics[width=.99\linewidth]{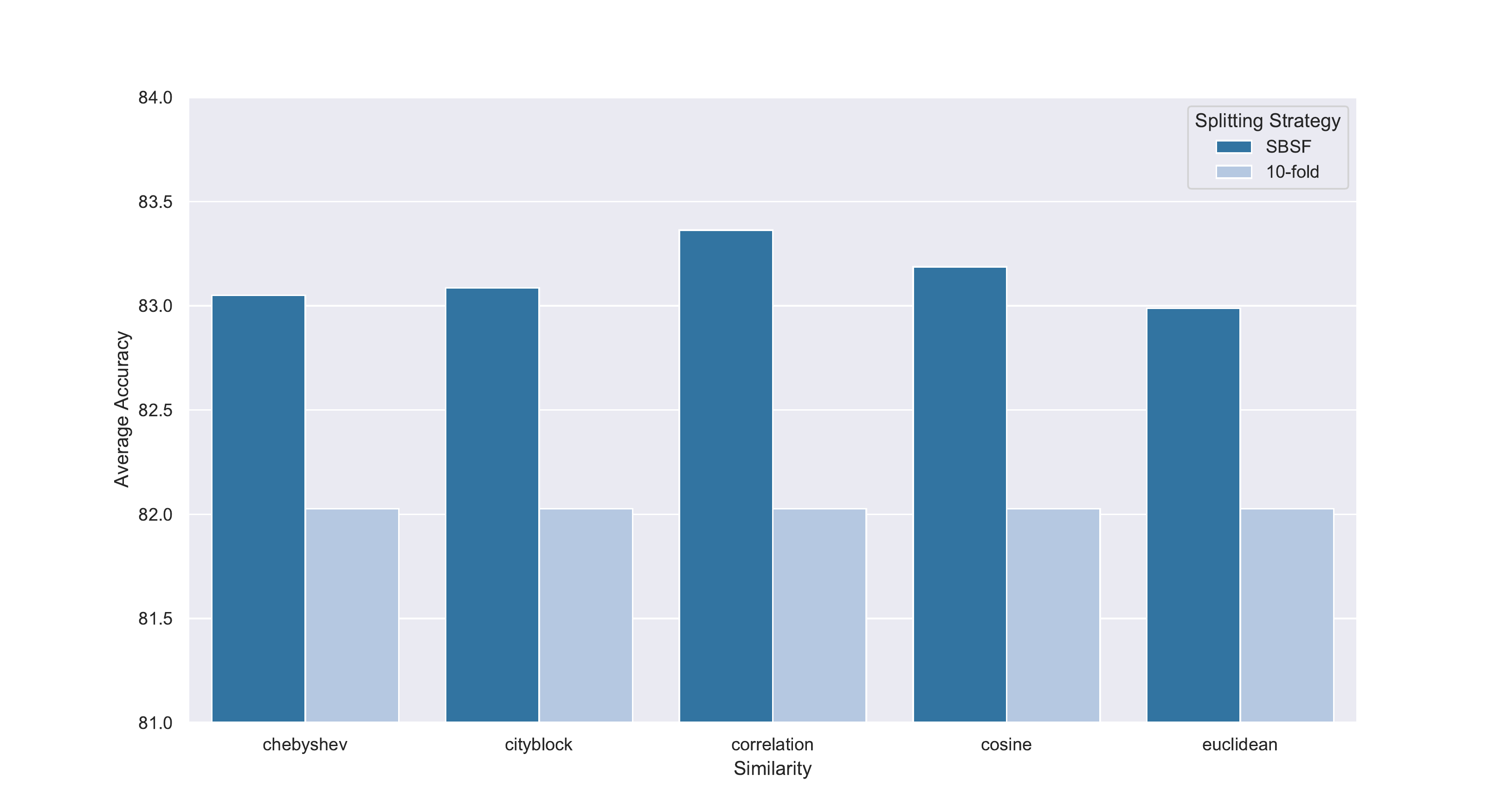}
	\caption{Averaged accuracy over datasets and models for each SBSF similarity compared to 10-fold strategy.}
	\label{fig:acc_split_strategy}
\end{figure}

The averaged accuracy for all datasets and models for each similarity compared to the 10-fold strategy is presented in Figure~\ref{fig:acc_split_strategy}. It is easy to notice that the Correlation presented the most remarkable performance among all similarities.

\begin{figure}[hbt!]
	\centering
		\includegraphics[width=.99\linewidth]{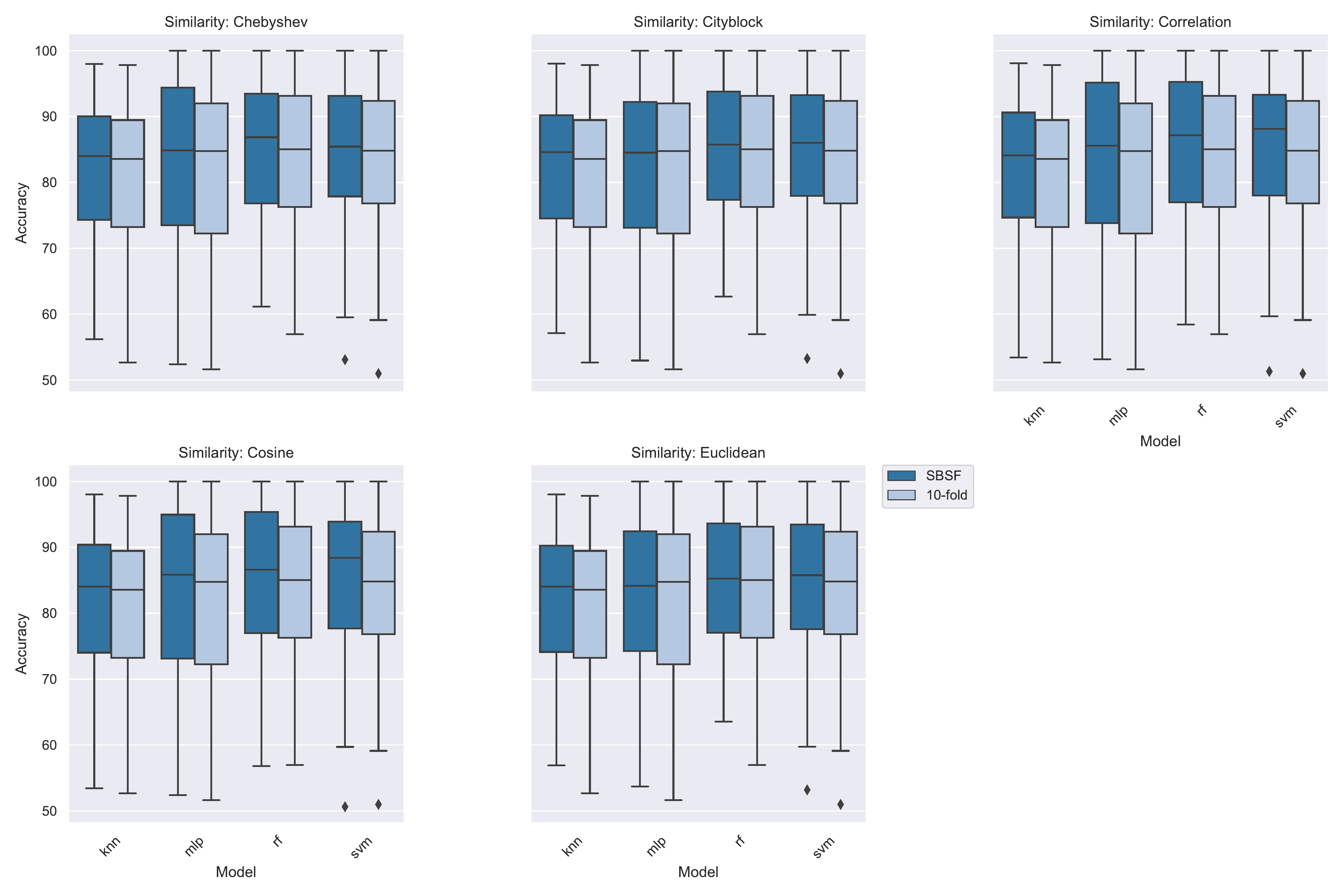}
	\caption{Boxplots with accuracies for each similarity and model of SBSF compared to 10-fold strategy.}
	\label{fig:boxplots_accuracies}
\end{figure}

The boxplots comparing absolute accuracy for each similarity and model, regarding all datasets are shown in Figure~\ref{fig:boxplots_accuracies}. The median values in SBSF with Correlation similarity was always more significant than the 10-fold strategy for every model. The same occurs with other similarity/model scenarios.

\begin{figure}[hbt!]
	\centering
		\includegraphics[width=.99\linewidth]{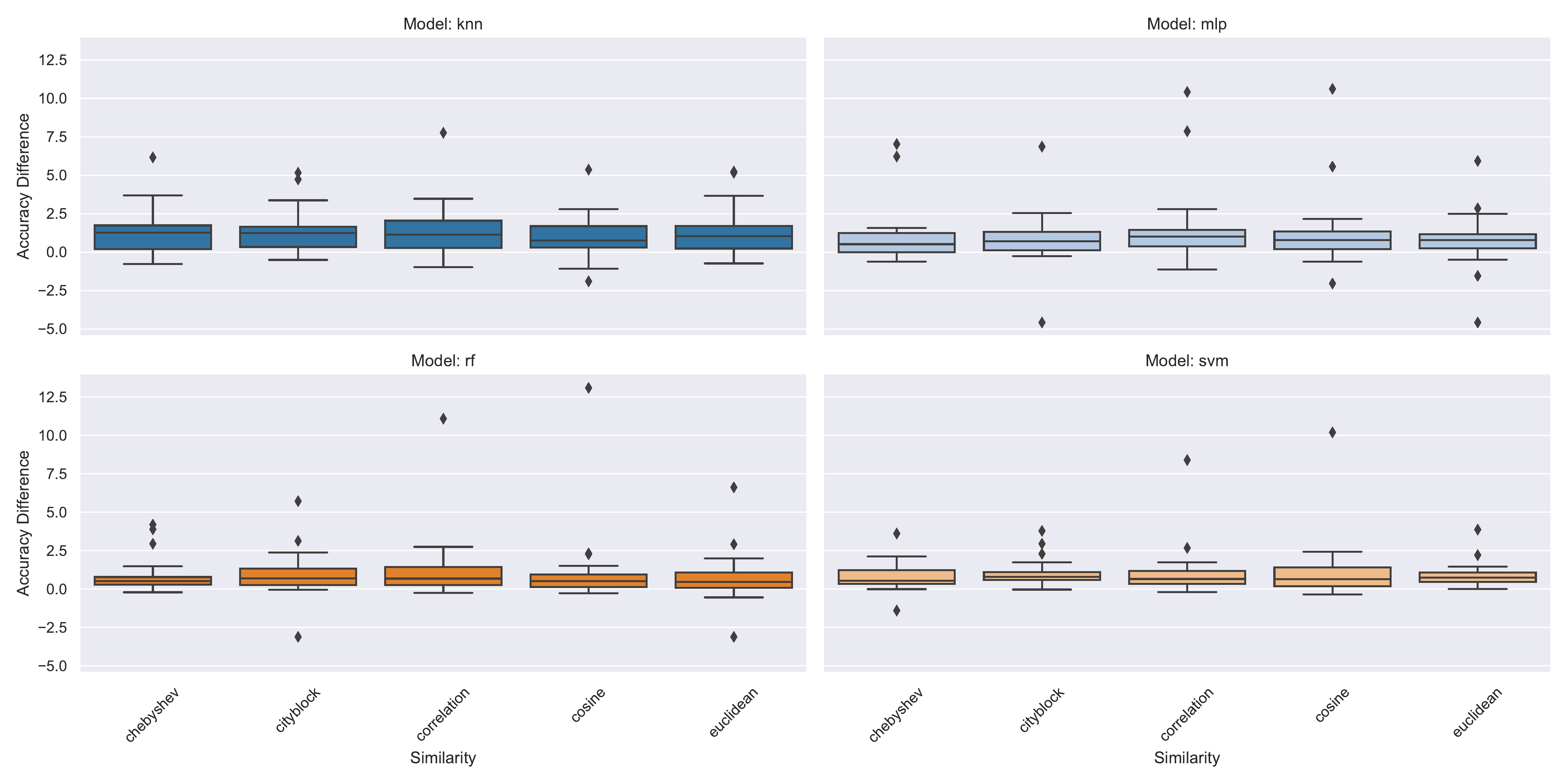}
	\caption{Boxplots with accuracies differences SBSF - 10-fold.}
	\label{fig:acc_differences}
\end{figure}

In Figure~\ref{fig:acc_differences}, the boxplots depict the differences between the accuracies of the SBSF and 10-fold. In general, the similarities had few negative differences, with some of them treated as outliers.

In Table~\ref{table:diff_train_test}, we can see that the average difference between the accuracy in the training - test set for SBSF is less than the 10-fold. It may indicate that the training and test set distributions are closer in the SBSF strategy than ordinary 10-fold. It can also indicate that SBSF intrinsically regularizes the training as the more significant this difference, the greater the chance of overfitting.

\begin{table}[width=.45\linewidth,cols=2]
\caption{Averaged difference of accuracy between train and test sets for each splitting strategy.}\label{table:diff_train_test}
\begin{tabular*}{\tblwidth}{@{} LC@{} }
    \toprule
\textbf{Strategy}           & \textbf{Difference Train-Test}   \\
\midrule
\textbf{SBSF + Chebyshev}   & 7.442                            \\
\textbf{SBSF + Cityblock}   & 7.292                            \\
\textbf{SBSF + Correlation} & 7.173                            \\
\textbf{SBSF + Cosine}      & 7.192                            \\
\textbf{SBSF + Euclidean}   & 7.397                            \\
\textbf{10-fold}            & 8.103                          \\
\bottomrule
\end{tabular*}
\end{table}

In Table~\ref{table:model_similarity_scores}, the scores of SBSF vs. 10-fold are presented for each model and each similarity. The MLP is probably the most sensible technique to its hyperparameters between the assessed models. Specifically the learning rate, number of neurons, and number of epochs, which we have fixed as 0.001, 20, and 300, respectively. As we have assessed this simple architecture regardless of the dataset, which has several numbers of samples/features, this probably led to the most significant number of losses without any hyperparameter exploration. The similarity Cityblock appears to be the best similarity since it got the highest number of wins (79.55\%). However, due to the highest accuracy of Correlation similarity shown in Table~\ref{table:similarities_accuracies}, we have chosen the Correlation similarity with 75\% of wins, 22.73\% of ties, and only 2.27\% of losses with an average increase in test set accuracy of 1.336\% for the next analysis. The Correlation similarity probably had better performances because the vectors' mean of $u$ and $v$ are subtracted, which led to smaller magnitudes of the vectors, facilitating the calculation of L2-norms better-extracting similarities information among the samples.

\begin{table}[width=.99\linewidth,cols=16]
\caption{Number of losses, ties and wins regarding the models and similarities of the SBSF over 10-fold splitting.}\label{table:model_similarity_scores}
\begin{tabular*}{\tblwidth}{@{} CCCCCCCCCCCCCCCC@{} }
    \toprule
    \multirow{2}{*}{\textbf{Model}}           & \multicolumn{3}{c}{\textbf{Chebyshev}}      & \multicolumn{3}{c}{\textbf{Cityblock}}      & \multicolumn{3}{c}{\textbf{Euclidean}}      & \multicolumn{3}{c}{\textbf{Cosine}}         & \multicolumn{3}{c}{\textbf{Correlation}}    \\
               \cline{2-4} \cline{5-7} \cline{8-10} \cline{11-13} \cline{14-16}
               & \textbf{loss} & \textbf{tie} & \textbf{win} & \textbf{loss} & \textbf{tie} & \textbf{win} & \textbf{loss} & \textbf{tie} & \textbf{win} & \textbf{loss} & \textbf{tie} & \textbf{win} & \textbf{loss} & \textbf{tie} & \textbf{win} \\
    \midrule
\textbf{KNN}                    & 2             & 4            & 16           & 0             & 3            & 19           & 0             & 6            & 16           & 3             & 2            & 17           & 1             & 5            & 16           \\
\textbf{MLP}                    & 3             & 4            & 15           & 1             & 7            & 14           & 2             & 1            & 19           & 3             & 3            & 16           & 1             & 4            & 17           \\
\textbf{RF}                     & 0             & 5            & 17           & 0             & 4            & 18           & 0             & 6            & 16           & 1             & 8            & 13           & 0             & 5            & 17           \\
\textbf{SVM}                    & 1             & 5            & 16           & 0             & 3            & 19           & 0             & 5            & 17           & 1             & 5            & 16           & 0             & 6            & 16           \\ \midrule
\textbf{Total}                           & 6             & 18           & 64           & 1             & 17           & 70           & 2             & 18           & 68           & 8             & 18           & 62           & 2             & 20           & 66           \\
\textbf{\%}                              & 6.82        & 20.45      & 72.73      & 1.14        & 19.32      & 79.55      & 2.27        & 20.45      & 77.27      & 9.09        & 20.45      & 70.45      & 2.27        & 22.73      & 75     \\ \bottomrule
\end{tabular*}
\end{table}

In the following subsections, we present a detailed comparative analysis of each model using the Correlation similarity with SBSF and ordinary 10-fold.

\subsection{K-Nearest Neighbors}
The train and test accuracies with their respective differences (SBSF-10-fold) of SBSF with Correlation similarity in KNN, presented in Table~\ref{table:knn}, increased the test accuracy in 17 datasets, remaining the same in 4 and losing in 1 case when compared with 10-fold splitting according to Wilcoxon statistical test. An increase in the training set did not necessarily accompany the test set's increase with the same magnitude. For example, in the vehicle dataset, the test accuracy increased 2.071\% while the training stayed almost the same. The standard deviation also decreased in SBSF. The average difference of train-test accuracies of SBSF and 10-fold is $1-(87.028-81.222)/(86.322-79.791)=11.1\%$ lower than the ordinary 10-fold.

\begin{table}
\caption{Average accuracy for KNN with SBSF using Correlation similarity. Highlighted values are higher according to Wilcoxon Significance test.}\label{table:knn}
\begin{tabular*}{\tblwidth}{@{} LLLLLLL@{} }
\toprule
\multirow{2}{*}{\textbf{Dataset}} & \multicolumn{3}{c}{\textbf{Train}}  & \multicolumn{3}{c}{\textbf{Test}} \\ \cline{2-4} \cline{5-7}
                                  & \textbf{SBSF} & \textbf{10fold} & \textbf{Difference} & \textbf{SBSF} & \textbf{10fold} & \textbf{Difference} \\ \midrule
                             
\textbf{australian}       & \textbf{88.851 (0.447)} & 87.902 (0.614)          & 0.949 (-0.167)  & \textbf{86.397 (2.603)} & 84.536 (4.562)           & 1.861 (-1.959)  \\
\textbf{arcene}           & 92.38 (1.044)           & 92.178 (1.025)          & 0.202 (0.019)   & \textbf{88.526 (6.447)} & 85.05 (8.313)            & 3.476 (-1.866)  \\
\textbf{balance-scale}    & \textbf{89.789 (0.864)} & 86.99 (0.962)           & 2.799 (-0.098)  & \textbf{85.8 (2.952)}   & 82.624 (4.115)           & 3.176 (-1.163)  \\
\textbf{btsc}             & 81.862 (0.774)          & 81.827 (0.829)          & 0.035 (-0.055)  & 77.041 (2.751)          & 76.912 (3.787)           & 0.129 (-1.036)  \\
\textbf{car}              & \textbf{98.997 (0.211)} & 98.914 (0.257)          & 0.083 (-0.046)  & \textbf{96.661 (1.292)} & 95.672 (1.638)           & 0.989 (-0.346)  \\
\textbf{cmsc}             & \textbf{94.797 (0.321)} & 93.889 (0.419)          & 0.908 (-0.098)  & \textbf{94.113 (1.771)} & 92.815 (2.277)           & 1.298 (-0.506)  \\
\textbf{cnae-9}           & \textbf{91.249 (0.746)} & 91.118 (0.718)          & 0.131 (0.028)   & \textbf{84.759 (3.019)} & 84.185 (2.875)           & 0.574 (0.144)   \\
\textbf{credit-g}         & 81.686 (0.58)           & 81.653 (0.682)          & 0.033 (-0.102)  & \textbf{74.34 (3.131)}  & 74.04 (3.914)            & 1.3 (-0.783)    \\
\textbf{dbworld-bodies}   & \textbf{66.022 (2.165)} & 60.118 (2.712)          & 5.904 (-0.547)  & \textbf{62.6 (6.114)}   & 54.833 (5.858)           & 7.767 (0.256)   \\
\textbf{diabetes}         & \textbf{82.386 (0.655)} & 82.079 (0.7)            & 0.307 (-0.045)  & \textbf{75.658 (3.738)} & 73.841 (4.877)           & 1.817 (-1.139)  \\
\textbf{hill-valley}      & 72.873 (0.679)          & 72.833 (0.763)          & 0.04 (-0.084)   & \textbf{53.442 (3.784)} & 52.673 (4.487)           & 0.769 (-0.703)  \\
\textbf{ilpd}             & \textbf{79.23 (1.005)}  & 78.237 (0.931)          & 0.993 (0.074)   & \textbf{67.123 (4.262)} & 65.147 (4.379)           & 1.976 (-0.117)  \\
\textbf{ionosphere}       & 87.245 (0.582)          & \textbf{87.809 (0.748)} & -0.564 (-0.166) & 84.5 (4.427)            & 85.211 (5.436)           & -0.711 (-1.009) \\
\textbf{libras move}     & \textbf{87.093 (1.046)} & 85.006 (1.199)          & 2.087 (-0.153)  & \textbf{77.967 (6.01)}  & 75.083 (7.749)           & 2.884 (-1.739)  \\
\textbf{lsvt}             & \textbf{90.528 (1.877)} & 89.021 (1.491)          & 1.507 (0.386)   & 82.667 (10.143)         & \textbf{83.647 (10.737)} & -0.98 (-0.594)  \\
\textbf{madelon}          & 73.864 (0.656)          & 73.963 (0.592)          & -0.099 (0.064)  & 56.865 (3.122)          & 57.135 (3.106)           & -0.27 (0.016)   \\
\textbf{musk}             & \textbf{98.865 (0.069)} & 98.842 (0.083)          & 0.023 (-0.014)  & \textbf{98.102 (0.42)}  & 97.825 (0.593)           & 0.277 (-0.173)  \\
\textbf{satimage}         & \textbf{94.055 (0.141)} & 93.877 (0.159)          & 0.178 (-0.018)  & \textbf{91.309 (0.877)} & 90.88 (1.08)             & 0.429 (-0.203)  \\
\textbf{tic-tac-toe}      & 85.021 (0.67)           & 84.932 (0.688)          & 0.089 (-0.018)  & 83.758 (3.596)          & 83.466 (3.344)           & 0.292 (0.252)   \\
\textbf{vehicle}          & 82.02 (0.792)           & 82.025 (0.753)          & -0.005 (0.039)  & \textbf{72.183 (3.848)} & 69.788 (4.101)           & 2.395 (-0.253)  \\
\textbf{vowel}            & 98.053 (0.259)          & 98.038 (0.306)          & 0.015 (-0.047)  & \textbf{96.263 (1.834)} & 94.192 (2.459)           & 2.071 (-0.625)  \\
\textbf{wdbc}             & 97.742 (0.284)          & \textbf{97.825 (0.294)} & -0.083 (-0.01)  & 96.821 (2.141)          & 96.854 (2.452)           & -0.033 (-0.311) \\
\midrule
\textbf{Average}          & 87.028 (0.721)          & 86.322 (0.769)          & 0.706 (-0.048)  & 81.222 (3.558)          & 79.791 (4.188)           & 1.431 (-0.63)   \\
\textbf{Losses/Ties/Wins} & 2L/8T/12W               &                         &                 & 1L/5T/16W               &                          &                 \\                                                          
\bottomrule
\end{tabular*}
\end{table}

\subsection{Multi-layer Perceptron}
Table~\ref{table:mlp} shows the accuracies of train and test set with their differences of SBSF with Correlation similarity in MLP. The test accuracy increased in 17 cases, persisted the same in 4, and lost in 1. As in KNN, the difference in test and training accuracy was not proportional. For example, in dbworld-bodies dataset, the test accuracy increased 10.424\% while the training accuracy only increased 1.423\%. The standard deviation of the accuracies also had a large decrease. As we have not used a validation set to stop the training or choose the best MLP model, we can realize that the difference between accuracies obtained in the training and testing sets of 10-fold is higher than in SBSF. It is probably a sign of over-fitting in 10-fold splitting since the training error is much lower than the test error. We believe that, as we have better sampled the dataset through SBSF, a better knowledge extraction was performed. Thus, SBSF can act as an intrinsic regularizer difficulting the over-fitting, as the train-test accuracy difference of SBSF in this experiment is, on average, $1-(85.958-81.8)/(85.53-80.253)=21.205\%$ lower than the ordinary 10-fold.

\begin{table}
\caption{Average accuracy for MLP with SBSF using Correlation similarity. Highlighted values are higher according to Wilcoxon Significance test.}\label{table:mlp}
\begin{tabular*}{\tblwidth}{@{} LLLLLLL@{} }
\toprule
\multirow{2}{*}{\textbf{Dataset}} & \multicolumn{3}{c}{\textbf{Train}}  & \multicolumn{3}{c}{\textbf{Test}} \\ \cline{2-4} \cline{5-7}
                                  & \textbf{SBSF} & \textbf{10fold} & \textbf{Difference} & \textbf{SBSF} & \textbf{10fold} & \textbf{Difference} \\ \midrule
\textbf{australian}     & 89.98 (0.932)             &  90.188 (1.053)           &  -0.208 (-0.121)  & \textbf{88.0 (2.292) }    &  86.652 (4.127)           &  1.348 (-1.835)    \\
\textbf{arcene}         & 85.175 (22.52)            &  84.122 (22.788)          &  1.053 (-0.268)   & \textbf{74.895 (18.009) } &  72.1 (16.379)            &  2.795 (1.63)      \\
\textbf{balance-scale}  & \textbf{97.409 (0.896)}  &  97.031 (1.053)           &  0.378 (-0.157)   & \textbf{96.5 (2.276) }    &  95.406 (2.682)           &  1.094 (-0.406)    \\
\textbf{btsc}           & \textbf{80.763 (0.707)}  &  79.979 (0.769)           &  0.784 (-0.062)   & \textbf{80.23 (2.273) }   &  78.943 (3.204)           &  1.287 (-0.931)    \\
\textbf{car}            &  97.394 (0.92)             &  97.222 (1.021)           &  0.172 (-0.101)   & \textbf{96.503 (1.537) }  &  96.134 (1.83)            &  0.369 (-0.293)    \\
\textbf{cmsc}           & \textbf{99.73 (0.345) }   &  99.265 (0.643)           &  0.465 (-0.298)   & \textbf{96.132 (2.185) }  &  94.722 (2.718)           &  1.41 (-0.533)     \\
\textbf{cnae-9}         & \textbf{99.897 (0.045) }  &  99.889 (0.046)           &  0.008 (-0.001)   & \textbf{92.287 (2.158) }  &  91.898 (2.279)           &  0.389 (-0.121)    \\
\textbf{credit-g}       &  87.772 (1.943)            &  87.996 (1.846)           &  -0.224 (0.097)   & \textbf{73.66 (3.781) }   &  72.74 (4.031)            &  0.92 (-0.25)      \\
\textbf{dbworld-bodies} & \textbf{100.0 (0.0) }     &  98.577 (0.695)           &  1.423 (-0.695)   & \textbf{99.4 (1.897) }    &  88.976 (12.706)          &  10.424 (-10.809)  \\
\textbf{diabetes}       & \textbf{79.863 (0.736) }  &  79.674 (0.964)           &  0.189 (-0.228)   &  77.276 (3.709)            &  76.886 (4.195)           &  0.39 (-0.486)     \\
\textbf{hill-valley}    &  63.855 (4.764)            &  64.449 (5.4)             &  -0.594 (-0.636)  &  62.725 (5.963)            &  63.848 (7.206)           &  -1.123 (-1.243)   \\
\textbf{ilpd}           & \textbf{75.774 (1.069) }  &  74.704 (0.928)           &  1.07 (0.141)     & \textbf{73.211 (4.147) }  &  71.496 (4.169)           &  1.715 (-0.022)    \\
\textbf{ionosphere}     &  99.15 (0.387)             &  99.161 (0.374)           &  -0.011 (0.013)   &  91.912 (4.341)            &  92.048 (4.345)           &  -0.136 (-0.004)   \\
\textbf{libras move}    & \textbf{66.163 (11.839) } &  60.639 (10.532)          &  5.524 (1.307)    & \textbf{60.533 (12.558) } &  52.667 (10.987)          &  7.866 (1.571)     \\
\textbf{lsvt}           &  99.972 (0.062)            &  99.991 (0.028)           &  -0.019 (0.034)   & \textbf{86.917 (8.747) }  &  85.462 (9.689)           &  1.455 (-0.942)    \\
\textbf{madelon}        &  59.112 (10.967)           &  59.407 (11.629)          &  -0.295 (-0.662)  & \textbf{53.546 (4.572) }  &  53.023 (4.181)           &  0.523 (0.391)     \\
\textbf{musk}           &  99.997 (0.009)            &  99.997 (0.008)           &  0.0 (0.001)      & \textbf{99.997 (0.01) }   &  99.986 (0.03)            &  0.011 (-0.02)     \\
\textbf{satimage}       & \textbf{85.368 (1.583) }  &  85.015 (1.53)            &  0.353 (0.053)    & \textbf{84.611 (1.787) }  &  84.062 (1.64)            &  0.549 (0.147)     \\
\textbf{tic-tac-toe}    &  91.989 (2.371)            &  92.149 (2.525)           &  -0.16 (-0.154)   &  86.579 (3.627)            &  86.285 (4.408)           &  0.294 (-0.781)    \\
\textbf{vehicle}        &  77.016 (2.689)            &  76.824 (2.159)           &  0.192 (0.53)     & \textbf{74.39 (4.133) }   &  73.099 (4.117)           &  1.291 (0.016)     \\
\textbf{vowel}          &  55.879 (7.081)            & \textbf{56.663 (7.298) } &  -0.784 (-0.217)  & \textbf{53.152 (7.265) }  &  51.626 (8.279)           &  1.526 (-1.014)    \\
\textbf{wdbc}           &  98.813 (0.302)            &  98.719 (0.32)            &  0.094 (-0.018)   &  97.143 (1.98)             & \textbf{ 97.505 (1.993) } &  -0.362 (-0.013)   \\
\midrule
\textbf{Average}        &  85.958 (3.28)             &  85.53 (3.346)            &  0.428 (-0.066)   &  81.8 (4.511)              &  80.253 (5.236)           &  1.547 (-0.725)   \\
\textbf{Losses/Ties/Wins} & 1L/12T/9W                                                         &                                                                   &                                                           & 1L/4T/17W                                                         &                                                                    &\\                                                          
\bottomrule
\end{tabular*}
\end{table}

\subsection{Support Vector Machine}
Regarding SVM in Table~\ref{table:svm}, the usage of SBSF also increased the test set accuracy in 16 datasets, remaining the same in 6 and with no losses compared with 10-fold splitting. One can observe an increase in the test set performance without not necessarily having an increase in the training set, as it was the case with previous classifiers. For example, in the arcene dataset, the test accuracy increased $1.734\%$ while presenting a statistically significant decrease in the training set, which was 0.283 smaller than with 10-fold. The SBSF strategy performed worse in 4 cases regarding the training set (arcene, car, credit-g, and ionosphere) while showing statistically significant increases in the same datasets at the test set, except in ionosphere. The train-test accuracy difference of SBSF in this experiment is, on average, $1-(89.458-84.38)/(89.029-83.259)=11.993\%$ lower than the ordinary 10-fold.

\begin{table}
\caption{Average accuracy for SVM with SBSF using Correlation similarity. Highlighted values are higher according to Wilcoxon Significance test.}\label{table:svm}
\begin{tabular*}{\tblwidth}{@{} LLLLLLL@{} }
\toprule
\multirow{2}{*}{\textbf{Dataset}} & \multicolumn{3}{c}{\textbf{Train}}  & \multicolumn{3}{c}{\textbf{Test}} \\ \cline{2-4} \cline{5-7}
                                  & \textbf{SBSF} & \textbf{10fold} & \textbf{Difference} & \textbf{SBSF} & \textbf{10fold} & \textbf{Difference} \\ \midrule
\textbf{australian}        & \textbf{87.843 (0.347)} & 87.626 (0.567)          & 0.217 (-0.22)   & \textbf{86.368 (2.152)} & 85.449 (3.993)  & 0.919 (-1.841)  \\
\textbf{arcene}            & 92.573 (0.895)          & \textbf{92.856 (1.09)}  & -0.283 (-0.195) & \textbf{78.684 (6.468)} & 76.95 (8.617)   & 1.734 (-2.149)  \\
\textbf{balance-scale}     & \textbf{92.689 (0.253)} & 91.561 (0.354)          & 1.128 (-0.101)  & \textbf{92.133 (1.141)} & 90.464 (1.546)  & 1.669 (-0.405)  \\
\textbf{btsc}              & \textbf{78.03 (0.206)}  & 77.594 (0.405)          & 0.436 (-0.199)  & \textbf{77.5 (0.872)}   & 76.781 (1.827)  & 0.719 (-0.955)  \\
\textbf{car}               & 97.973 (0.202)          & \textbf{98.13 (0.16)}   & -0.157 (0.042)  & 96.778 (1.173) & 96.846 (1.312)  & -0.068 (-0.139) \\
\textbf{cmsc}              & \textbf{97.57 (0.298)}  & 96.864 (0.379)          & 0.706 (-0.081)  & \textbf{93.566 (1.271)} & 92.722 (1.723)  & 0.844 (-0.452)  \\
\textbf{cnae-9}            & 99.313 (0.126)          & 99.298 (0.146)          & 0.015 (-0.02)   & \textbf{91.852 (2.351)} & 91.389 (2.624)  & 0.463 (-0.273)  \\
\textbf{credit-g}          & 82.409 (0.504)          & \textbf{82.567 (0.541)} & -0.158 (-0.037) & \textbf{77.11 (2.853)}  & 76.13 (3.204)   & 0.98 (-0.351)   \\
\textbf{dbworld-bodies}    & \textbf{100.0 (0.0)}    & 98.577 (0.695)          & 1.423 (-0.695)  & \textbf{92.6 (9.962)}   & 84.214 (15.234) & 8.386 (-5.272)  \\
\textbf{diabetes}          & \textbf{80.553 (0.521)} & 80.233 (0.555)          & 0.32 (-0.034)   & \textbf{78.184 (3.807)} & 76.979 (4.407)  & 1.205 (-0.6)    \\
\textbf{hill-valley}       & \textbf{53.554 (0.403)} & 53.295 (0.486)          & 0.259 (-0.083)  & \textbf{51.317 (2.686)} & 50.982 (3.326)  & 0.335 (-0.64)   \\
\textbf{ilpd}              & \textbf{71.93 (0.0)}    & 71.355 (0.084)          & 0.575 (-0.084)  & \textbf{71.93 (0.0)}    & 71.356 (0.762)  & 0.574 (-0.762)  \\
\textbf{ionosphere}        & 95.918 (0.417)          & \textbf{96.075 (0.466)} & -0.157 (-0.049) & 93.588 (2.622)          & 93.394 (4.226)  & 0.194 (-1.604)  \\
\textbf{libras move}       & \textbf{91.511 (1.284)} & 89.917 (0.84)           & 1.594 (0.444)   & \textbf{82.867 (5.571)} & 81.306 (5.883)  & 1.561 (-0.312)  \\
\textbf{lsvt}              & \textbf{88.796 (1.07)}  & 87.187 (1.573)          & 1.609 (-0.503)  & 83.0 (9.131)            & 82.558 (9.329)  & 0.442 (-0.198)  \\
\textbf{madelon}           & 95.768 (0.207)          & 95.767 (0.249)          & 0.001 (-0.042)  & \textbf{59.681 (2.677)} & 59.115 (2.859)  & 0.566 (-0.182)  \\
\textbf{musk}              & 100.0 (0.0)             & 100.0 (0.0)             & 0.0 (0.0)       & 100.0 (0.0)             & 100.0 (0.0)     & 0.0 (0.0)       \\
\textbf{satimage}          & \textbf{91.148 (0.127)} & 90.951 (0.149)          & 0.197 (-0.022)  & \textbf{90.172 (0.938)} & 89.879 (1.083)  & 0.293 (-0.145)  \\
\textbf{tic-tac-toe}       & 92.91 (0.464)           & 92.994 (0.388)          & -0.084 (0.076)  & 89.884 (2.31)           & 89.54 (2.963)   & 0.344 (-0.653)  \\
\textbf{vehicle}           & \textbf{83.381 (0.671)} & 81.655 (0.757)          & 1.726 (-0.086)  & \textbf{77.963 (3.11)}  & 75.297 (4.098)  & 2.666 (-0.988)  \\
\textbf{vowel}             & 95.807 (0.406)          & 95.872 (0.417)          & -0.065 (-0.011) & \textbf{93.717 (2.015)} & 92.687 (2.271)  & 1.03 (-0.256)   \\
\textbf{wdbc}              & \textbf{98.395 (0.222)} & 98.258 (0.23)           & 0.137 (-0.008)  & 97.464 (2.04)           & 97.663 (1.97)   & -0.199 (0.07)   \\ \midrule
\textbf{Average}           & 89.458 (0.392)          & 89.029 (0.479)          & 0.429 (-0.087)  & 84.38 (2.961)           & 83.259 (3.784)  & 1.121 (-0.823) \\
\textbf{Losses/Ties/Wins}  & 4L/5T/13W               &                         &                 & 0L/6T/16W               &                 &\\                                                          
\bottomrule
\end{tabular*}
\end{table}

%%%%%%%%%%%%%%%%%%%%%%%%%%%%%%%%%%%%%%%
% RANDOM FOREST RESULTS
%%%%%%%%%%%%%%%%%%%%%%%%%%%%%%%%%%%%%%%
\subsection{Random Forest}
Assessing the impact of SBSF to RF, we can see from Table~\ref{table:rf} that the test set accuracy increased in 17 datasets, remaining the same in 5 and with no losses regarding the 10-fold splitting. The corresponding increase behavior in the training and testing sets of previous classifiers also applies to RF. The train-test accuracy difference of SBSF in this experiment is, on average, $1-(99.701-86.051)/(99.637-84.803)=7.982\%$ lower than the ordinary 10-fold.

\begin{table}
\caption{Average accuracy for RF with SBSF using Correlation similarity. Highlighted values are higher according to Wilcoxon Significance test.}\label{table:rf}
\begin{tabular*}{\tblwidth}{@{} LLLLLLL@{} }
\toprule
\multirow{2}{*}{\textbf{Dataset}} & \multicolumn{3}{c}{\textbf{Train}}  & \multicolumn{3}{c}{\textbf{Test}} \\ \cline{2-4} \cline{5-7}
                                  & \textbf{SBSF} & \textbf{10fold} & \textbf{Difference} & \textbf{SBSF} & \textbf{10fold} & \textbf{Difference} \\ \midrule
\textbf{australian}     & 100.0 (0.0)          & 99.998 (0.005) & 0.002 (-0.005) & \textbf{88.574 (2.773)} & 87.101 (3.817)  & 1.473 (-1.044)  \\
\textbf{arcene}         & 100.0 (0.0)          & 100.0 (0.0)    & 0.0 (0.0)      & \textbf{85.263 (6.047)} & 83.0 (8.066)    & 2.263 (-2.019)  \\
\textbf{balance-scale}  & 100.0 (0.0)          & 100.0 (0.0)    & 0.0 (0.0)      & \textbf{85.7 (2.874)}   & 82.96 (3.31)    & 2.74 (-0.436)   \\
\textbf{btsc}           & 93.435 (0.228)       & 93.434 (0.321) & 0.001 (-0.093) & \textbf{74.743 (3.332)} & 73.959 (4.652)  & 0.784 (-1.32)   \\
\textbf{car}            & 100.0 (0.0)          & 100.0 (0.0)    & 0.0 (0.0)      & \textbf{98.737 (0.835)} & 98.472 (1.033)  & 0.265 (-0.198)  \\
\textbf{cmsc}           & 99.996 (0.013)       & 100.0 (0.0)    & -0.004 (0.013) & \textbf{93.283 (1.063)} & 92.574 (1.584)  & 0.709 (-0.521)  \\
\textbf{cnae-9}         & 100.0 (0.0)          & 100.0 (0.0)    & 0.0 (0.0)      & \textbf{93.167 (2.235)} & 92.611 (2.221)  & 0.556 (0.014)   \\
\textbf{credit-g}       & 100.0 (0.0)          & 100.0 (0.0)    & 0.0 (0.0)      & \textbf{76.96 (3.019)}  & 76.25 (3.845)   & 0.71 (-0.826)   \\
\textbf{dbworld-bodies} & \textbf{100.0 (0.0)} & 98.577 (0.695) & 1.423 (-0.695) & \textbf{97.2 (5.583)}   & 86.119 (15.418) & 11.081 (-9.835) \\
\textbf{diabetes}       & 99.999 (0.005)       & 100.0 (0.0)    & -0.001 (0.005) & \textbf{76.987 (4.363)} & 76.352 (4.249)  & 0.635 (0.114)   \\
\textbf{hill-valley}    & 100.0 (0.0)          & 100.0 (0.0)    & 0.0 (0.0)      & \textbf{58.442 (4.273)} & 56.955 (3.93)   & 1.487 (0.343)   \\
\textbf{ilpd}           & 100.0 (0.0)          & 99.998 (0.006) & 0.002 (-0.006) & \textbf{72.246 (3.949)} & 70.509 (4.883)  & 1.737 (-0.934)  \\
\textbf{ionosphere}     & 100.0 (0.0)          & 100.0 (0.0)    & 0.0 (0.0)      & 93.059 (3.5)            & 93.309 (4.202)  & -0.25 (-0.702)  \\
\textbf{libras\_move}   & 99.996 (0.012)       & 100.0 (0.0)    & -0.004 (0.012) & \textbf{84.367 (6.373)} & 83.333 (6.231)  & 1.034 (0.142)   \\
\textbf{lsvt}           & 100.0 (0.0)          & 100.0 (0.0)    & 0.0 (0.0)      & \textbf{85.25 (8.986)}  & 83.968 (9.292)  & 1.282 (-0.306)  \\
\textbf{madelon}        & 100.0 (0.0)          & 100.0 (0.0)    & 0.0 (0.0)      & 71.492 (2.538)          & 71.615 (2.929)  & -0.123 (-0.391) \\
\textbf{musk}           & 100.0 (0.0)          & 100.0 (0.0)    & 0.0 (0.0)      & \textbf{99.994 (0.019)} & 99.98 (0.046)   & 0.014 (-0.027)  \\
\textbf{satimage}       & 99.999 (0.001)       & 99.999 (0.002) & 0.0 (-0.001)   & \textbf{92.092 (0.79)}  & 91.712 (1.068)  & 0.38 (-0.278)   \\
\textbf{tic-tac-toe}    & 100.0 (0.0)          & 100.0 (0.0)    & 0.0 (0.0)      & 95.947 (2.034)          & 95.919 (1.967)  & 0.028 (0.067)   \\
\textbf{vehicle}        & 100.0 (0.0)          & 100.0 (0.0)    & 0.0 (0.0)      & 75.5 (3.478)            & 75.251 (3.452)  & 0.249 (0.026)   \\
\textbf{vowel}          & 100.0 (0.0)          & 100.0 (0.0)    & 0.0 (0.0)      & \textbf{97.99 (1.476)}  & 97.515 (1.557)  & 0.475 (-0.081)  \\
\textbf{wdbc}           & 100.0 (0.0)          & 100.0 (0.0)    & 0.0 (0.0)      & 96.125 (2.335)          & 96.204 (2.303)  & -0.079 (0.032)  \\ \midrule
\textbf{Average}        & 99.701 (0.012)       & 99.637 (0.047) & 0.064 (-0.035) & 86.051 (3.267)          & 84.803 (4.093)  & 1.248 (-0.826) \\
\textbf{Losses/Ties/Wins}  & 0L/21T/1W               &                         &                 & 0L/5T/17W               &                 &\\                                                          
\bottomrule
\end{tabular*}
\end{table}

% The most improved datasets present a small number of features, ranging from 4 to 12. Except for satimage, the datsets with 20+ features such as credit-g, ionosphere, and wdbc did not improve with SBSF. It probably happens due to our similarity function (Euclidean distance), which tends to perform poorly when subjected to high dimensional data as the distance to the nearest sample approaches the distance of the farthest sample \cite{Beyer1999}. Accuracy in satimage, which has 36 features, probably increased with SBSF due to the larger number of samples, reducing the impact of Euclidean distance problem in high dimensional spaces.

The test accuracy of each classifier applied to each dataset using SBSF with Correlation similarity is summarized in Table~\ref{table:all_classifiers}. The RF presented better results in 8 of 22 cases. Since Decision Trees are used internally to divide spaces, and SBSF acts improving these spatial representations, most improvements were achieved.

\begin{table}
\caption{Average test accuracy for all classifiers using Correlation similarity with SBSF. Best absolute accuracy for each dataset are highlighted.}\label{table:all_classifiers}
\begin{tabular*}{\tblwidth}{@{} LLLLL@{} }
\toprule
\textbf{Dataset} & \textbf{RF}                                                       & \textbf{MLP}                                                      & \textbf{KNN}                                                      & \textbf{SVM}                                                      \\ \midrule
\textbf{australian}                  & \textbf{88.574 (2.773)} & 88.0 (2.292)            & 86.397 (2.603)          & 86.368 (2.152)          \\
\textbf{arcene}                      & 85.263 (6.047)          & 74.895 (18.009)         & \textbf{88.526 (6.447)} & 78.684 (6.468)          \\
\textbf{balance-scale}               & 85.7 (2.874)            & \textbf{96.5 (2.276)}   & 85.8 (2.952)            & 92.133 (1.141)          \\
\textbf{btsc}                        & 74.743 (3.332)          & \textbf{80.23 (2.273)}  & 77.041 (2.751)          & 77.5 (0.872)            \\
\textbf{car}                         & \textbf{98.737 (0.835)} & 96.503 (1.537)          & 96.661 (1.292)          & 96.778 (1.173)          \\
\textbf{cmsc}                        & 93.283 (1.063)          & \textbf{96.132 (2.185)} & 94.113 (1.771)          & 93.566 (1.271)          \\
\textbf{cnae-9}                      & \textbf{93.167 (2.235)} & 92.287 (2.158)          & 84.759 (3.019)          & 91.852 (2.351)          \\
\textbf{credit-g}                    & 76.96 (3.019)           & 73.66 (3.781)           & 74.34 (3.131)           & \textbf{77.11 (2.853)}  \\
\textbf{dbworld-bodies}              & 97.2 (5.583)            & \textbf{99.4 (1.897)}   & 62.6 (6.114)            & 92.6 (9.962)            \\
\textbf{diabetes}                    & 76.987 (4.363)          & 77.276 (3.709)          & 75.658 (3.738)          & \textbf{78.184 (3.807)} \\
\textbf{hill-valley}                 & 58.442 (4.273)          & \textbf{62.725 (5.963)} & 53.442 (3.784)          & 51.317 (2.686)          \\
\textbf{ilpd}                        & 72.246 (3.949)          & \textbf{73.211 (4.147)} & 67.123 (4.262)          & 71.93 (0.0)             \\
\textbf{ionosphere}                  & 93.059 (3.5)            & 91.912 (4.341)          & 84.5 (4.427)            & \textbf{93.588 (2.622)} \\
\textbf{Libras move}                 & \textbf{84.367 (6.373)} & 60.533 (12.558)         & 77.967 (6.01)           & 82.867 (5.571)          \\
\textbf{lsvt}                        & 85.25 (8.986)           & \textbf{86.917 (8.747)} & 82.667 (10.143)         & 83.0 (9.131)            \\
\textbf{madelon}                     & \textbf{71.492 (2.538)} & 53.546 (4.572)          & 56.865 (3.122)          & 59.681 (2.677)          \\
\textbf{musk}                        & 99.994 (0.019)          & 99.997 (0.01)           & 98.102 (0.42)           & \textbf{100.0 (0.0)}    \\
\textbf{satimage}                    & \textbf{92.092 (0.79)}  & 84.611 (1.787)          & 91.309 (0.877)          & 90.172 (0.938)          \\
\textbf{tic-tac-toe}                 & \textbf{95.947 (2.034)} & 86.579 (3.627)          & 83.758 (3.596)          & 89.884 (2.31)           \\
\textbf{vehicle}                     & 75.5 (3.478)            & 74.39 (4.133)           & 72.183 (3.848)          & \textbf{77.963 (3.11)}  \\
\textbf{vowel}                       & \textbf{97.99 (1.476)}  & 53.152 (7.265)          & 96.263 (1.834)          & 93.717 (2.015)          \\
\textbf{wdbc}                        & 96.125 (2.335)          & 97.143 (1.98)           & 96.821 (2.141)          & \textbf{97.464 (2.04)}  \\ \midrule
\textbf{Average}                     & \textbf{86.051 (3.267)} & 81.8 (4.511)            & 81.222 (3.558)          & 84.38 (2.961)           \\
\textbf{Score}                                 & \textbf{8}                                                                 & 7                                                                 & 1                                                                 & 6                                                                \\
\bottomrule
\end{tabular*}
\end{table}

Considering MLP, SVM, KNN and RF techniques and all the similarities scores, SBSF significantly increased the test accuracy in 330 cases (75\%), remained statistically similar in 91 (20.68\%), and decreased in 19 cases (4.32\%).

\section{Conclusion}
We believe that every model, including research and/or industrial applications, could benefit from this strategy to prepare the data to be learned, generating better models with increased real-world usage performance. After the model is trained, the real-world data to be presented should follow approximately the same distribution of the prepared training set. If this happens, probably the model could increase its performance if it uses the SBSS.

The SBSF showed statistically significant performance increases in the test set in 75\% of the cases. We believe that the a-posteriori distribution of the input space regions can be better explored, and this information can be incorporated into the model through a careful data splitting process used during the training phase. It is a low-cost strategy to increase models' performance by only changing how the training data is presented. We expect this approach to deploying models in the academy and industry scenarios with better performances. 

As future works, we intend to evaluate SBSS in other classifiers, including investigating their hyperparameters. We also plan to assess the proposal with different splitting strategies, such as Holdout, with and without stratification. An analysis of regression models with SBSS is also relevant. Other similarities functions may also benefit the SBSS strategy. Also, we intend to use the average of similarities instead of the sum in the algorithm that may benefit the performance.

% \appendix
% \section{My Appendix}
% Appendix sections are coded under \verb+\appendix+.

% \verb+\printcredits+ command is used after appendix sections to list 
% author credit taxonomy contribution roles tagged using \verb+\credit+ 
% in frontmatter.


\bibliographystyle{elsarticle-num}
%\bibliographystyle{cas-model2-names}

\bibliography{cas-refs}

\newpage
\bio{}
Felipe C. Farias received the B.Sc. in Computer Engineering in 2014, in this occasion he awarded the Academic Laurea of Engineering, the M.Sc. degree in Computer Intelligence in 2016, both from University of Pernambuco (UPE). Ph.D. student of Machine Learning at Federal University of Pernambuco (UFPE). He is currently a Professor at Federal Institute of Education, Science and Technology of Pernambuco (IFPE). His research interests include Neural Networks, Deep Learning, Machine Learning, Network Science. More information at https://scholar.google.com/citations?hl=pt-BR\&user=zwRHO14AAAAJ
http://lattes.cnpq.br/4598958786544738
\endbio

% \bio{figs/pic1}
\bio{}
Teresa B. Ludermir received the Ph.D. degree in Artificial Neural Networks in 1990 from Imperial College, University of London, UK. From 1991 to 1992, she was a lecturer at Kings College London. She joined the Center of Informatics at Federal University of Pernambuco,  Brazil, in September 1992, where she is currently a Professor and head of the Computational Intelligence Group. She has published over 300 articles in scientific journals and conferences, three books in Neural Networks (in Portuguese) and organized three of the Brazilian Symposium on Neural Networks. Her research interests include weightless Neural Networks, hybrid neural systems, machine learning and applications of Neural Networks. She is an IEEE senior member and a Research fellow level 1A (the higher level) of the National Research Council of Brazil (CNPq). More information at https://scholar.google.com.br/citations?hl=pt-BR\&user=w-tKJOwAAAAJ
https://orcid.org/0000-0002-8980-6742
http://lattes.cnpq.br/6321179168854922
\endbio

\bio{}
Prof. Carmelo J. A. Bastos-Filho was born in Recife, Brazil, in 1978. He received the B.Sc. in electronics engineering and the M.Sc. and Ph.D. degrees in electrical engineering from Federal University of Pernambuco (UFPE) in 2000, 2003, and 2005, respectively. In 2006, he received the best Brazilian thesis award in electrical engineering. His interests are related to: the development of protocols and algorithms to manage and to design optical communication networks, development of novel swarm intelligence algorithms and the deployment of swarm intelligence for complex optimization and clustering problems, development and application of multiobjective optimization and many-objective optimization for real-world problems, Developement and application of soft deep learning techniques, development of solutions using network sciences for big data, applications in robotics and applications in biomedical problems. He is currently an Associate Professor at the Polytechnic School of the University of Pernambuco. He was the scientist-in-chief of the technological Park for Electronics and Industry 4.0 of Pernambuco. He is currently the director for science parks and graduate studies of Pernambuco. He is an IEEE senior member and a Research fellow level 1D of the National Research Council of Brazil (CNPq). He published over 230 full papers in journals and conferences and advised over 50 PhD and MSc candidates. More information at http://scholar.google.com/citations?user=t3A96agAAAAJ\&hl=en
\endbio
\end{document}